\documentclass{article}
\usepackage{spconf,amsmath,graphicx}

\usepackage{times}  
\usepackage{helvet}  
\usepackage{courier}  
\usepackage{url}  
\usepackage{graphicx}  
\usepackage{color} 
\usepackage{comment}
\usepackage{subfigure}

\title{Deep Learning for Vertex Reconstruction of Neutrino-Nucleus Interaction Events
with Combined Energy and Time Data}

\name{\it Linghao Song$^\dagger$, Fan Chen$^\dagger$, Steven R. Young$^\ddagger$, 
Catherine D. Schuman$^\ddagger$, \\ \it
Gabriel Perdue$^\perp$, and Thomas E. Potok$^\ddagger$}
\address{
$^\dagger$Duke University, Durham, North Carolina, 27708\\
$^\ddagger$Oak Ridge National Laboratory, Oak Ridge, Tennessee, 37831\\
$^\perp$Fermi National Accelerator Laboratory, Batavia, Illinois, 60510
}

\begin{document}
\maketitle

\begin{abstract}
We present a deep learning approach for vertex reconstruction of neutrino-nucleus interaction events, a problem in the domain of high energy physics.  In this approach, we combine both energy and timing data that are collected in the MINERvA detector to perform classification and regression tasks.  We show that the resulting network achieves higher accuracy than previous results while requiring a smaller model size and less training time. In particular, the proposed model outperforms  the  state-of-the-art by 4.00\% on  classification accuracy. For the regression task, our model achieves 0.9919 on the coefficient of determination, higher than the previous work (0.96). 

\let\thefootnote\relax\footnote{Notice: This manuscript has been authored by UT-Battelle, LLC under contract DE-AC05-00OR22725, and Fermi Research Alliance, LLC (FRA) under contract DE-AC02-07CH11359 with the US Department of Energy (DOE). The US government retains and the publisher, by accepting the article for publication, acknowledges that the US government retains a nonexclusive, paid-up, irrevocable, worldwide license to publish or reproduce the published form of this manuscript, or allow others to do so, for US government purposes. DOE will provide public access to these results of federally sponsored research in accordance with the DOE Public Access Plan (\url{http://energy.gov/downloads/doe-public-access-plan}).}
\end{abstract}
\vspace{-1.25em}

\section{Introduction}
MINERvA (Main Injuector Experiment for v-A)~\cite{ALIAGA2014130} is a leading-edge program at Fermi National Accelerator Laboratory.
The primary focus of the MINERvA experiment is to understand neutrino properties and reactions.  Neutrinos are subatomic particles that rarely interact with normal matter as they only interact via weak subatomic force and gravity and they have extremely small mass.  
The study of neutrinos may help physicists understand the matter-antimatter imbalance in the universe \cite{Acciarri:2015uup}. 
However, understanding their interactions with nuclear matter poses significant challenges: they probe aspects of nuclear structure that are not accessible with electrons, photons, or protons \cite{Mosel:2016cwa}.

In the MINERvA experiment, the detector is exposed to the Neutrinos at the Main Injector (NuMI) neutrino beam \cite{adamson2016numi}.  
The detector records both energy and timing information that can be used to determine where neutrino-nucleus interaction events occur.  
Precise determination of the interaction vertex, also known as vertex reconstruction \cite{terwilliger2017vertex}, is required to identify the target nucleus in MINERvA.

\begin{figure}[htb!]
\centering
\vspace{-5pt}
\includegraphics[width=0.8 \columnwidth]{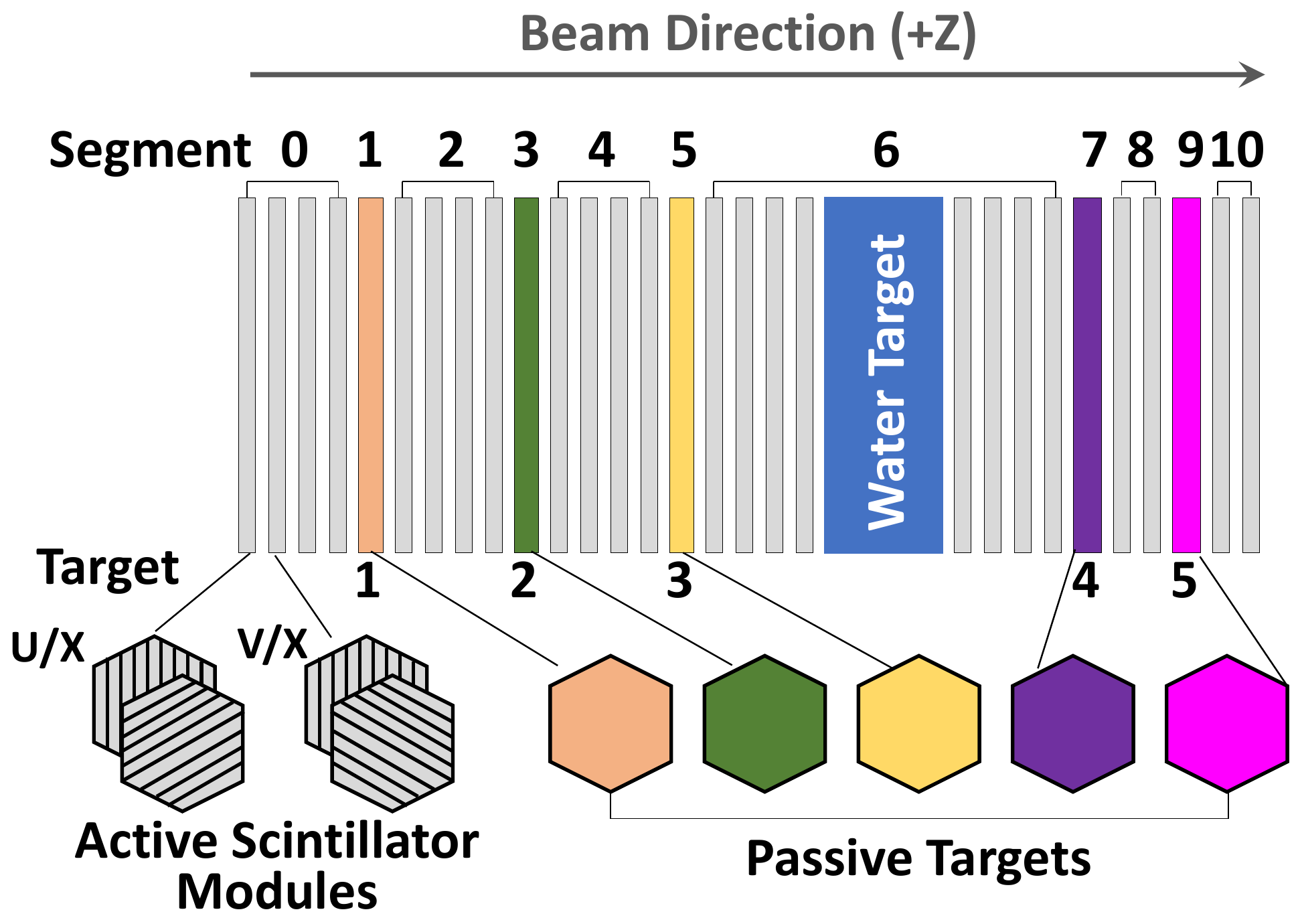}
\vspace{-5pt}
\caption{Simple illustration of the MINERvA detector.}
\label{fig_target}
\vspace{-5pt}
\end{figure}

Fig.\ref{fig_target} illustrates a simple detector layout.
We eliminate details of the detector and the physics measurements obtained, but readers can refer to~\cite{ALIAGA2014130,terwilliger2017vertex} for details.  
The core of the detector consists of a series of alternating active and passive target regions along the beam direction. 
The passive targets are solid layers of different materials or their combinations, e.g., carbon, iron, lead, and tanks of liquid helium and water. Note in the datasets considered in this work, the liquid/water target is empty.
The active targets are plastic scintillator (a
hydrocarbon) modules. 
Each active module contains a pair of planes with scintillator strips aligned in one of three orientations: X, U or V.
Strips in X planes are oriented vertically and U and V strips are oriented $\pm60^\circ$ relative to X.
Each module contains either a U or V plane, followed by an X, such that the pattern is interleaved “UXVXUXVX,” etc.
Energy and timing values collected from the detector are mapped to pixel values in an image, which can be used for subsequently vertex reconstruction.

A key issue associated with data from this experiment and scientific data in general is that it is often extremely difficult to obtain labels. For example, there may be only a handful of experts in the world capable of labeling experimental data effectively, and even then, it may be impossible to establish ground-truth labels for the data that multiple experts will agree is correct. As such, much of the scientific data that can be used for training is generated using simulations.  
For this study, millions of simulated neutrino-nucleus scattering
events were created and represented as images.
In this case, deep learning approaches to analyze the data can help physicists in quickly interpreting results from the experiments. 


The contributions of this work are:
(1) the incorporation of both energy and time lattices in one network to boost classification accuracy, 
(2) the utilization of transfer learning to improve performance on regression of absolute position, and
(3) a new network topology to combine three views (X, U, V) and reduce model size.

\section{Data Description} 

{ The dataset used for training, validation and testing consisted of $1,453,884$ simulated events.
Neutrino-nucleus interactions were simulated using the GENIE Neutrino Monte Carlo Generator \cite{andreopoulos2015genie}, and the propagation of the resulting radiation through the bulk detector was simulated using the Geant4 toolkit \cite{Agostinelli:2002hh}.}
For each event, there is both an energy lattice and a time lattice, each of which consists of three views: an X-view, a U-view, and a V-view. 
The images from the X-view are $127\times94$ pixels, while the others (U-view and V-view) are $127\times47$ pixels.  
Each pixel in the energy lattice gives information about the average energy value over the detection event at that point, while each pixel in the time lattice recorded the timing information in nanoseconds relative to when the interaction is predicted to occur.

\begin{figure}[htb!]
\centering
\vspace{-5pt}
\includegraphics[width=0.95 \columnwidth]{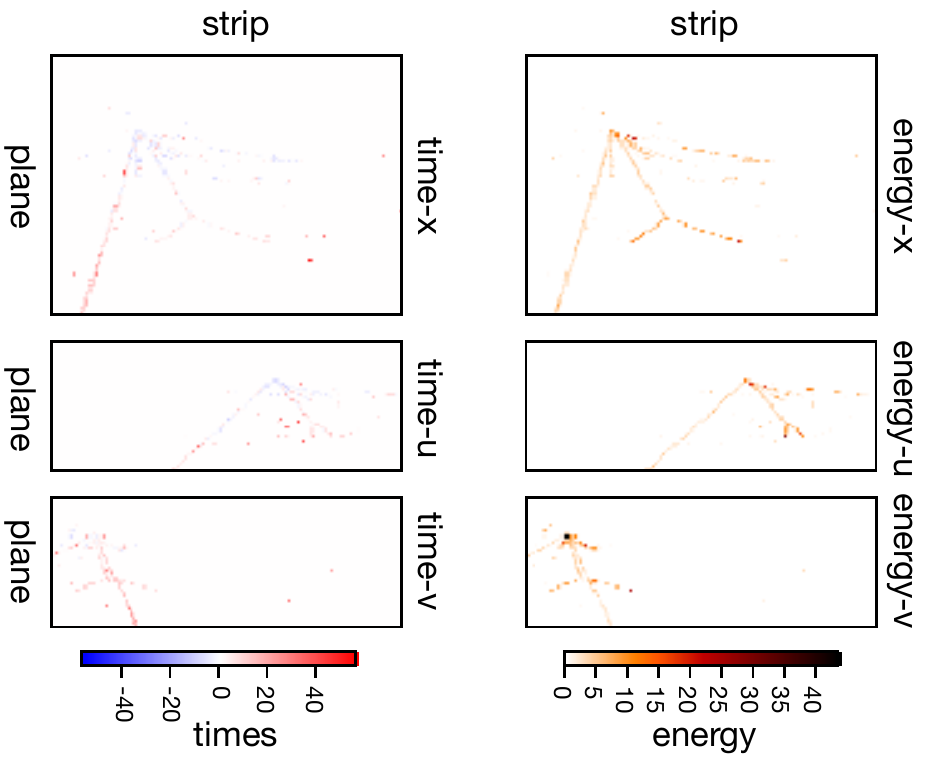}
\vspace{-10pt}
\caption{Example of the neutrino interaction events.}
\label{fig_evt2}
\vspace{-5pt}
\end{figure}

There are three scales at which we can attempt to predict the vertex location. The largest scale is a segment. The detector can be split into 11 segments, each of which consists of multiple planes within the detector. 
Approaching a smaller scale, the detector can be split into each of the planes.
{Planes are thin, horizontally stacked bundles of active sensors. They are oriented roughly perpendicular to the neutrino beam.}
Finally, the vertex location can be defined as the absolute measured position (Z) inside the detector.




\section{Previous Works}
The initial approach to vertex reconstruction for this dataset was to identify linear tracks and calculated the intersection points of multiple tracks as the vertex.  
This method fails for certain types of events; in particular, it is difficult to identify vertex when tracks are non-linear or differentiate individual tracks when the number of track is great. 
A previous work has applied deep learning (specifically convolutional neural networks) to the energy lattice of the data (as images) to improve classification accuracy \cite{terwilliger2017vertex}.  Another previous work has applied spiking neural networks to the vertex reconstruction problem using the time lattice only \cite{schuman2017neuromorphic}, achieving comparable results to the convolutional approach for a single view of the data, which indicated that the timing data includes information relevant to the vertex reconstruction problem as well. It is worth noting that both of these approaches utilize an older version of the dataset that used a reduce input size as compared with the dataset used here.
{Another work explored the use of Domain Adversarial Neural Networks \cite{JMLR:v17:15-239} for controlling physics modeling bias \cite{Perdue:2018ihs}.}
In this work, we seek to combine both the energy lattice and time lattice in a convolutional neural network implementation. The neural network model is designed to predict the segment and absolute position (Z) of the neutrino events.
 
\section{Approach}

\subsection{Model for Segment Classification} 
\begin{figure*}[htb!]
\centering
\vspace{-0pt}
\subfigure[Classification model]{ 
    \label{fig_net} 
    \includegraphics[width=1.5 \columnwidth]{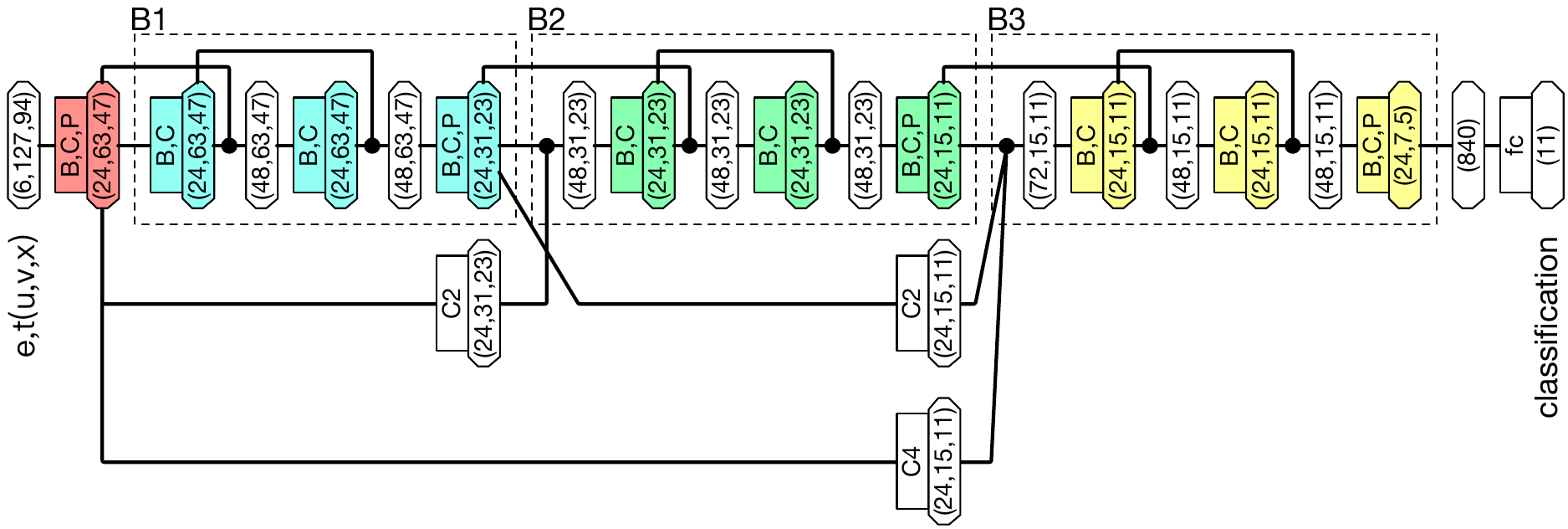}
} 
\hspace{0.25in} 
\subfigure[Regression model]{ 
    \label{fig_reg} 
    \includegraphics[width=0.33 \columnwidth]{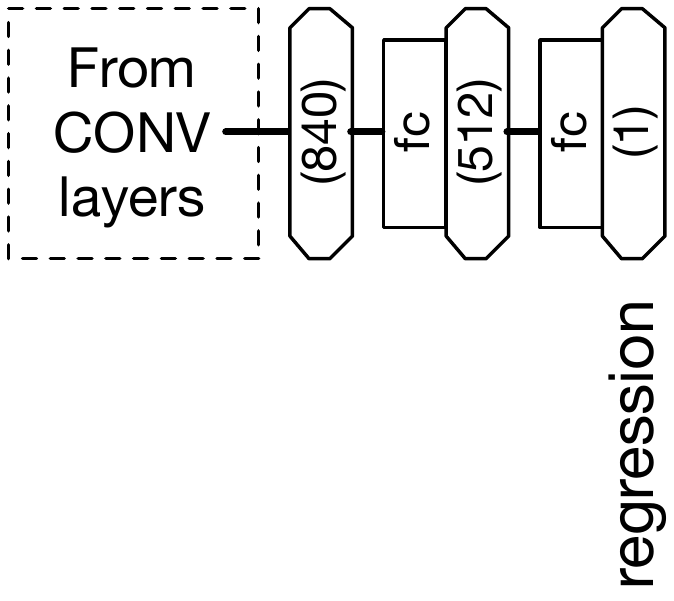}
}

\vspace{-5pt}
\caption{The networks for (a) segment classification and (b) Z regression. In the Z regression, the convolutional layers in the trained classification network were frozen, and two fully-connected layers were added and retrained.}
\vspace{-5pt}

\end{figure*}
To get a network with smaller size and to alleviate
the vanishing-gradient problem, inspired by the ResNet \cite{he2016deep} and DenseNet
\cite{huang2017densely}, we designed our network for segment classification as shown in Fig.\ref{fig_net}.

A rectangle
represents a series of operations. For example, a rectangle labeled as B,C,P
means that batch normalization (B),
convolution (C) and max pooling (P) are successively applied to the input tensors. All convolutions (C) 
in the network are configured with {\tt kernel\_size=3}, {\tt 
padding=1} and {\tt stride=1}, and a ReLU activation 
function is applied. 
The kernel size and stride for the max pooling (P) is 2. 
The octagon below a rectangle indicates the output tensor size, 
while an octagon on the flow indicates the concatenated
tensor size or the reformed tensor size. The tensor size 
format is $C,H,W$, where $C$ is the number of channels, 
$H$ is the height and $W$ is the width. For example, the
input tensor {\tt e,t(u,v,x)} has 2 groups (energy and 
timing) of 3 views (U, V, X), thus 6 channels of 127-by-94
matrix.

A black dot indicates the concatenation of two tensors 
by channel.
There are three blocks (B1, B2 and B3) in the network. Within each block, the input of each rectangle is a concatenation 
of two tensors: the output of a previous rectangle 
and a shortcut identical tensor. For each block, 
we also apply direct-connect from previous blocks.
For block B2, the direct-connect is a convolution (C2) with a 
kernel size and stride of 2. So, the input for B2 is the 
concatenation of two (24,31,23)-tensors, i.e., a tensor of size 
(48,31,23). For block B3, two direct-connects (C2 and C4) are
used. C4 is a convolution with kernel size and stride of 4. 
Thus, the input tensor size for block B3 is (72,15,11). Note that there
is no activation function for the convolutions of the three 
direct-connects.  For the final classification layer, we reformed the output 
tensor (24,7,5) to a tensor (840), and a fully-connected layer is employed.

\subsection{Model for Z Regression} 

Because we use the same data for Z-regression as we do for segment classification, it is quite natural to employ a transfer learning approach. 
Fig.\ref{fig_reg} shows the model used for Z regression. 
We use a well trained segment classification network, 
freeze all the convolutional layers, and add two fully-connected
layers (840-512-1) for regression.

\section{Experiments}

\subsection{Experimentation Details}
For the whole dataset ($1,453,884$ events), We separated this set into three parts, $1/9$ for testing, $1/9$ for validation, and $7/9$ for training. Each data sample contains 
three views (X, U, V) for timing and energy, 
i.e., a total of six views. The size of data for the X view 
is $127\times94$ while the size for U and V views is
$127\times47$. We repeated the U and V views on the second axis
to get a size of $127\times94$. Then, we concatenate the six $127\times94$ views to a obtain a tensor (6,127,94) as an input.
The original data was in a {\tt float32} format.  We first normalized the data by view and converted the data to {\tt uint8} format for fast training access. 
We also calculated the mean ($\mu$) and standard deviation ($\sigma$) on the training set, and applied whitening 
on input data; thus, the mean and standard deviation for all views are 0 and 1 respectively.

In the training of the classification network, we use
an SGD optimizer. The training takes 20 epochs. 
The learning rate is 0.1 for the first 10 epochs, 0.01 for the following 5 epochs 
and 0.001 for the last 5 epochs. 
SGD is configured with a momentum of 0.8 
and a weight decay of 5e-4, and the batch size is 256. Then we trained the regression 
network for 8 epochs with a learning rate of 0.001.
Two NVIDIA TITAN X (Pascal) GPUs are configured in data parallelism for training. The toolkit used is PyTorch. 

\subsection{Overall Results with Both Energy and Timing Data}

\begin{table}[htb]
\centering 
\vspace{-15pt}
\caption{Comparison to previous work}
\vspace{2pt}
\small
\begin{tabular}{||l|c|c|c||}
\hline
\textbf{}     & \textbf{\begin{tabular}[c]{@{}c@{}}\cite{terwilliger2017vertex}\end{tabular}} & \textbf{\begin{tabular}[c]{@{}c@{}}\cite{terwilliger2017vertex}*\end{tabular}} & \textbf{\begin{tabular}[c]{@{}c@{}}This Work\end{tabular}} 
\\ \hline
Image Size & $127\times 50$ & $127\times 94$ & $127\times 94$
\\ \hline
Accuracy      & 94.09\%                                                                                      & 88.97\%                                                                                      & \textbf{98.09\%}                                                               \\ \hline
$R^2$   & 0.96                                                                                         & 0.8886                                                                                            & \textbf{0.9919}                                                                \\ \hline
Model Size    & 14.5MB                                                                                       & -                                                                                            & \textbf{0.488 MB}                                                              \\ \hline
Training Time & 10 hrs                                                                                       & -                                                                                            & \textbf{2.5hrs}                                                                \\ \hline
\end{tabular}
\label{table_compre_to_previous}

\vspace{3pt}
\small{*These results were created by re-implementing the network from a previous work \cite{terwilliger2017vertex}, and evaluating against updated dataset.}
\vspace{-5pt}
\end{table}

We compare our work with a previous 
work \cite{terwilliger2017vertex} 
in Table \ref{table_compre_to_previous}.
For the segmentation classification, our model (shown in Fig.\ref{fig_net})
achieves an accuracy of 98.09\% on testing dataset, 4.00\% higher 
than that in \cite{terwilliger2017vertex}.
For the Z regression, the coefficient of determination
($R^2$) of our model (as shown in Fig.\ref{fig_reg}) is 0.9919, 
higher than the previous work (0.96). 
Additionally, the model size (the size of the trained model file) 
and the training time of our model are smaller than those 
of \cite{terwilliger2017vertex}.

\begin{figure}[htb!]
\centering
\vspace{-0pt}

\subfigure[Confusion matrix]{ 
    \includegraphics[width=0.44 \columnwidth]{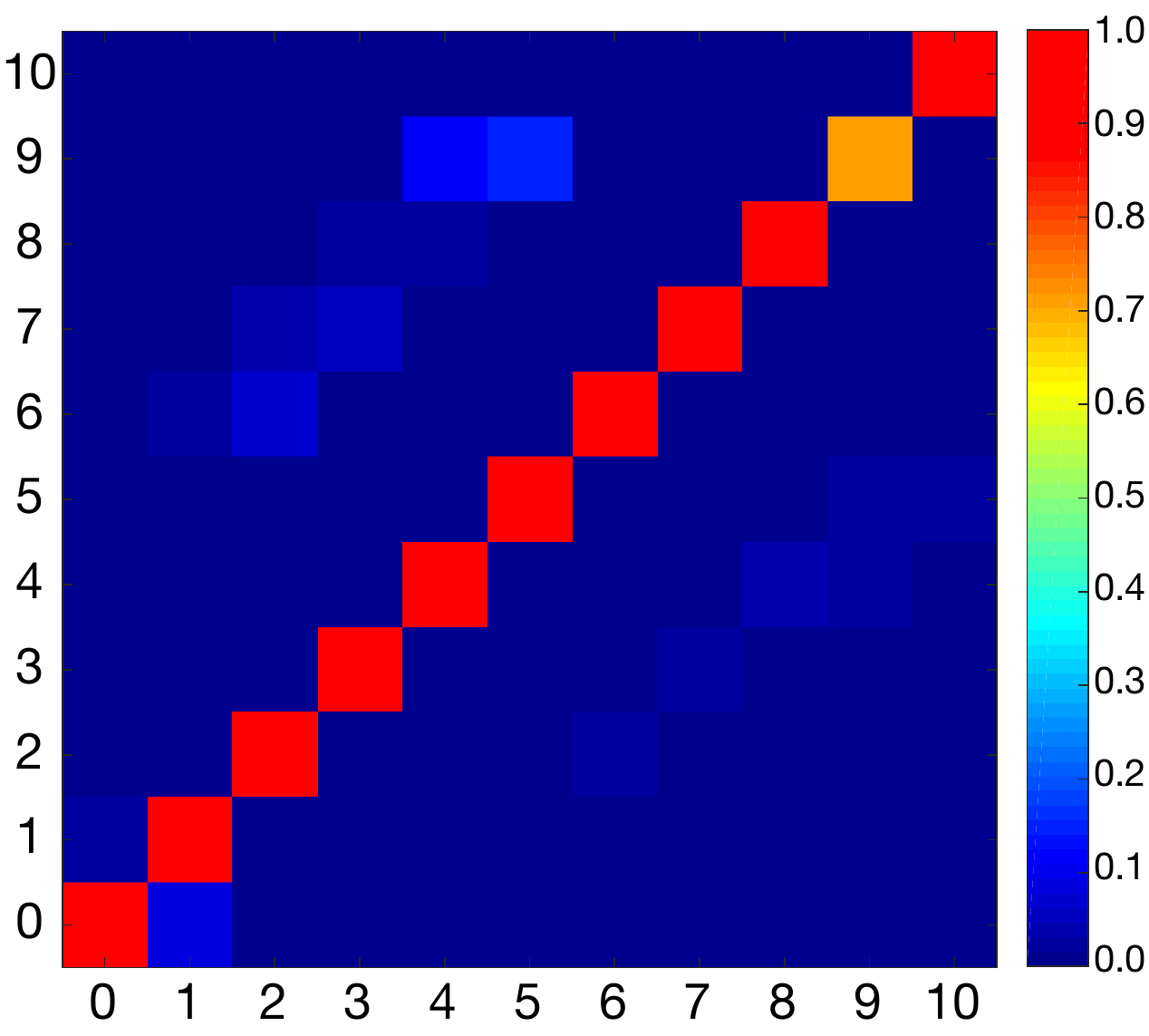}
    \label{fig_heatmap_et}
} 
\hspace{0.001in} 
\subfigure[Scatter plot]{ 
    \includegraphics[width=0.5 \columnwidth]{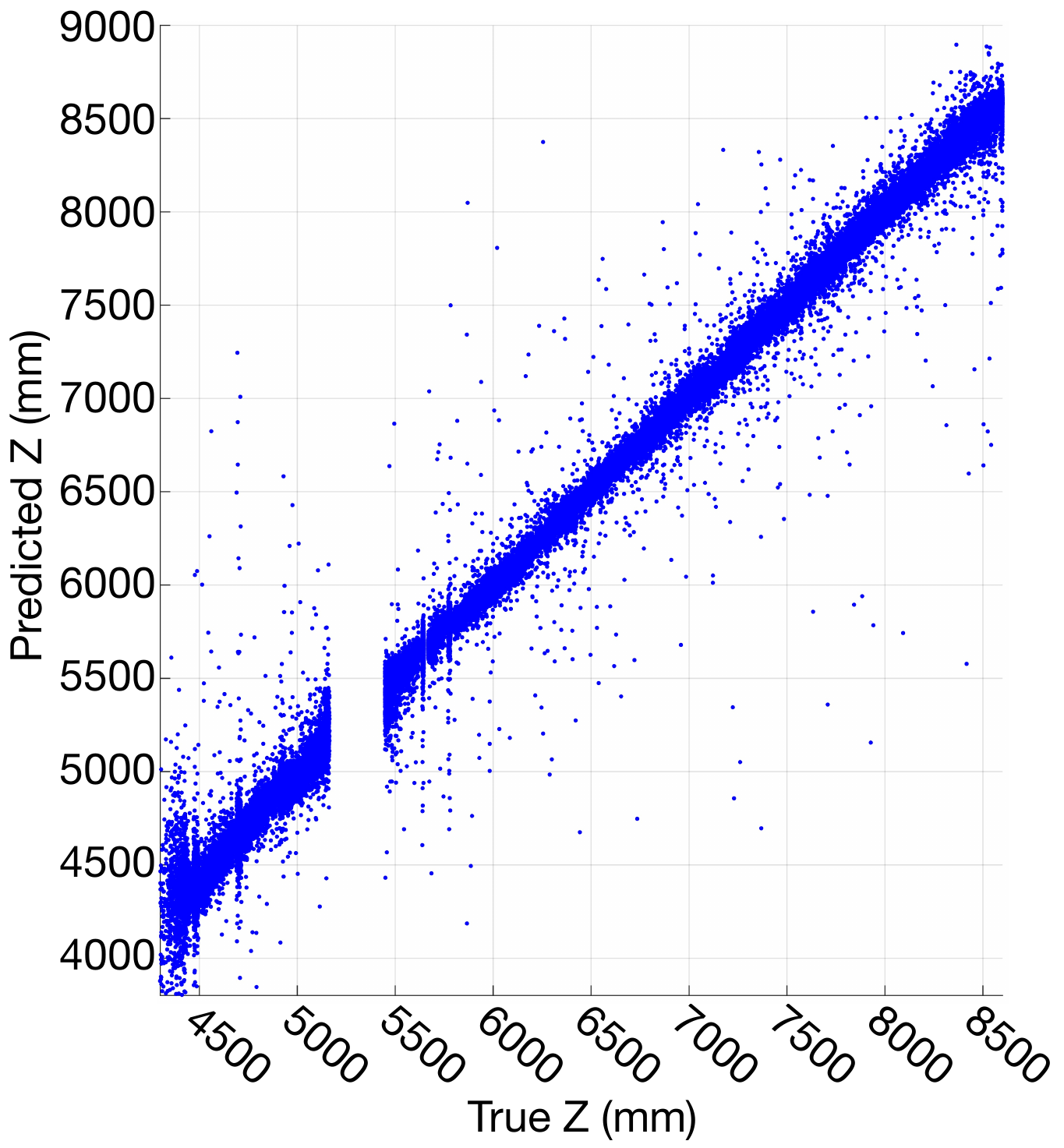}
    \label{fig_scatter}
}

\vspace{-10pt}
\caption{(a) Heatmap of confusion matrix of the segment classification and (b) scatter plot of the predicted and true Z (mm) on testing dataset with both energy and timing data. The water target(no detector) is around 5250mm, so there is a gap.}
\vspace{-10pt}
\end{figure}

Fig.\ref{fig_heatmap_et} shows the heatmap of the confusion matrix 
of the segment classification with both energy and timing data. 
We can see our model performs well for 
almost all segments except Segment 9.
We expect that this is due to the imbalance of the training data, in which only 0.47\% of the data is labeled as Segment 9.
We show the scatter plot of the predicted Z of
our regression model with both 
timing and energy data and true Z
in Fig.\ref{fig_scatter}. 
The standard deviation 
$\sigma$ of the difference of the predicted Z 
and the true Z is 115.61 mm.

\subsection{Classification and Regression with
Only Timing Data and Only Energy Data}

While we have both timing and energy data, we
are also interested in classification and regression with
only timing or energy data. In previous work
\cite{terwilliger2017vertex}, only energy data was used.

For segment classification with only timing or energy
data, we use the same network as shown in Fig.\ref{fig_net},
where the input is a tensor of X, U, V views of timing or
energy data, but not both simultaneously as in the previous section. Thus, the input tensor size is (3, 127, 94).
Table \ref{table_three_accuracy} shows the segment
classification accuracy of our model when both timing 
and energy data were used (combined) and when only timing 
or energy data was used. As shown in this table, when only timing 
or energy data was used, the accuracy is slightly degraded. 
However, it is clear that the energy data contributed more as the accuracy when only
energy data was used is only 0.13\% less than when both types of data are used.

\begin{table}[htb]
\centering 
\vspace{-12pt}
\caption{Segment classification accuracy on testing dataset}
\vspace{2pt}
\small
\begin{tabular}{||c|c|c||}
\hline 
Combined &Only Timing &Only Energy
\\ \hline 
\bf{98.09}\% & 96.92\% & 97.95\%
\\\hline 
\end{tabular}
\label{table_three_accuracy}
\vspace{-10pt}
\end{table}

\begin{table}[htb!]
\centering 
\vspace{-10pt}
\caption{Regression $R^2$ on testing dataset}
\vspace{2pt}
\small
\begin{tabular}{||c|c|c||}
\hline 
Combined &Only Timing &Only Energy
\\ \hline 
\bf{0.9919} & 0.9915 & 0.9901
\\\hline 
\end{tabular}
\label{table_three_R2}
\vspace{-5pt}
\end{table}

\begin{figure}[htb]
\centering
\vspace{-8pt}

\subfigure[Only timing data]{ 
    \includegraphics[width=0.36 \columnwidth]{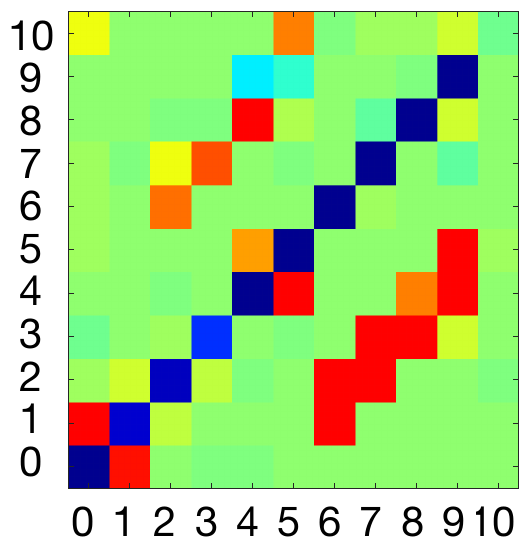}
} 
\hspace{0.1in} 
\subfigure[Only energy data]{ 
    \includegraphics[width=0.45 \columnwidth]{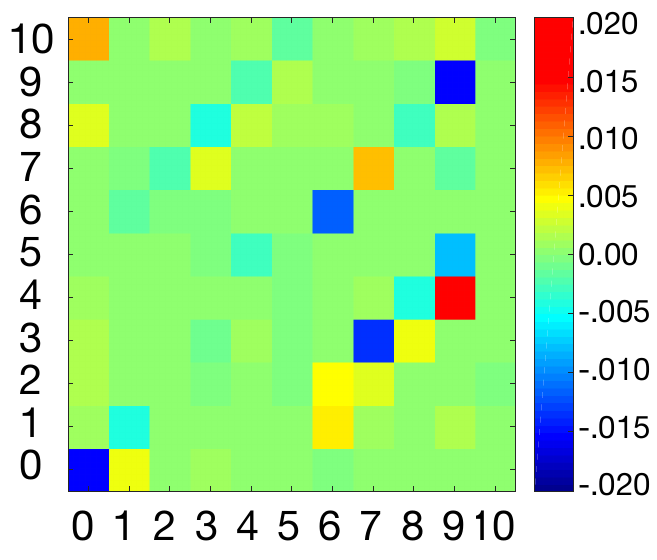}
}

\vspace{-10pt}
\caption{Heatmap of the difference of confusion matrix of
classifications on testing dataset with (a) only timing data 
and (b) only energy data compared to that 
when both timing and energy data were used.}
\label{fig_heatmap_e_and_t}
\vspace{-5pt}
\end{figure}

To get more details about the performance of the model when
only timing or energy data is used, we show the heat map of 
the difference of the confusion matrix when only timing 
or energy data was used, compared that when both were used in
Fig.\ref{fig_heatmap_e_and_t}. A warmer color square 
on the diagonal is better, while a warmer color square 
at other position than on the diagonal means misclassification.
Again, we find that energy data contributes more in the
segment classification.

Finally, we also compare the Z regression when only timing or 
energy data was used as shown in Table \ref{table_three_R2}. 
While the $R^2$ for the three scenarios are almost the same, but for a more accurate Z regression,
both timing and energy data should be used.

\section{Conclusion}
In this work we present a deep learning approach for vertex reconstruction in neutrino interaction data.  We demonstrate state-of-the-art results on this task, presenting a model that achieves higher accuracy on the dataset while also reducing both the training time required as well as the model size.  
For future work, we plan to explore the utilization of recurrent neural networks for capturing spatial-temporal features from within the event timing. 

\vspace{0.6em}
\noindent
{\bf Acknowledgements}


This material is based upon work supported by the U.S. Department of Energy, Office of Science, Office of Advanced Scientific Computing Research, Robinson Pino, program manager, under contract number DE-AC05-00OR22725.

We would like to thank the MINERvA collaboration for the use of their
simulated data and for many useful and stimulating conversations. MINERvA is
supported by the Fermi National Accelerator Laboratory under US Department of
Energy contract No. DE-AC02-07CH11359 which included the MINERvA construction
project. MINERvA construction support was also granted by the United States
National Science Foundation under Award PHY-0619727 and by the University of
Rochester. Support for participating MINERvA physicists was provided by NSF
and DOE (USA), by CAPES and CNPq (Brazil), by CoNaCyT (Mexico), by
CONICYT (Chile), by CONCYTEC, DGI-PUCP and IDI/IGIUNI (Peru), and by Latin
American Center for Physics (CLAF).

This research was supported in part by an appointment to the Oak Ridge National Laboratory ASTRO Program, sponsored by the U.S. Department of Energy and administered by the Oak Ridge Institute for Science and Education.

\bibliographystyle{IEEEbib}
\bibliography{main}

\end{document}